\documentclass[10pt,twocolumn,letterpaper]{article}

\usepackage[pagenumbers]{cvpr}  

\usepackage[accsupp]{axessibility}
\usepackage{capt-of}
\usepackage{xspace}
\usepackage{colortbl}
\usepackage{xcolor}
\usepackage{enumitem}
\usepackage{amsmath,amsthm,amssymb,amsfonts,dsfont,pifont,bm,bbm,mathrsfs,mathtools,nicefrac,extarrows,relsize}
\usepackage{algorithm,algpseudocode,listings}
\usepackage{booktabs,multirow,adjustbox,diagbox,threeparttable,tabularray,setspace}

\definecolor{cvprblue}{rgb}{0.21,0.49,0.74}
\usepackage[pagebackref,breaklinks,colorlinks,citecolor=cvprblue,bookmarks=false]{hyperref}
\usepackage{wrapfig}
\usepackage[capitalize]{cleveref}  

\crefname{section}{Sec.}{Secs.}
\Crefname{section}{Section}{Sections}
\crefname{appendix}{App.}{Apps.}
\Crefname{appendix}{Appendix}{Appendices}
\crefname{table}{Tab.}{Tabs.}
\Crefname{table}{Table}{Tables}
\crefname{figure}{Fig.}{Figs.}
\Crefname{figure}{Figure}{Figures}
\crefname{equation}{Eq.}{Eqs.}
\Crefname{equation}{Equation}{Equations}
\crefname{theorem}{Thm.}{Thms.}
\Crefname{theorem}{Theorem}{Theorems}
\crefname{lemma}{Lem.}{Lems.}
\Crefname{lemma}{Lemma}{Lemmas}
\crefname{remark}{Rem.}{Rems.}
\Crefname{remark}{Remark}{Remarks}
\crefname{corollary}{Cor.}{Cors.}
\Crefname{corollary}{Corollary}{Corollaries}
\crefname{algorithm}{Alg.}{Algs.}
\Crefname{algorithm}{Algorithm}{Algorithms}
\definecolor{cellred}{RGB}{213, 123, 101}
\definecolor{cellgreen}{RGB}{0, 205, 0}
\definecolor{cellblue}{RGB}{54, 125, 189}
\definecolor{codegreen}{rgb}{0,0.6,0}
\definecolor{codegray}{rgb}{0.5,0.5,0.5}
\definecolor{codepurple}{rgb}{0.58,0,0.82}
\definecolor{backcolour}{rgb}{1.0,1.0,1.0}
\definecolor{mygray}{gray}{0.92}
\lstdefinestyle{mystyle}{
    backgroundcolor=\color{backcolour},
    commentstyle=\color{codegreen},
    keywordstyle=\color{magenta},
    numberstyle=\tiny\color{codegray},
    stringstyle=\color{codepurple},
    basicstyle=\ttfamily\scriptsize,
    breakatwhitespace=false,
    breaklines=true,
    captionpos=b,
    keepspaces=true,
    numbers=left,
    numbersep=5pt,
    showspaces=false,
    showstringspaces=false,
    showtabs=false,
    tabsize=2
}
\newcommand{\tocite}[1]{{\color{red} [TO CITE]}}

\newcommand{\ours}[1]{Uni-AD}
\newcommand{\cmark}{\ding{51}}%
\newcommand{\xmark}{\ding{55}}%

\def\paperID{}
\def\confName{CVPR}
\def\confYear{2025}

\title{Contextual AD Narration with Interleaved Multimodal Sequence}

\author{Hanlin Wang$^{1,3}$ \quad Zhan Tong$^{2}$ \quad Kecheng Zheng$^{3}$ \quad Yujun Shen$^{3}$ \quad Limin Wang$^{1,4\dagger}$\\[0.3cm] $^1$State Key Laboratory for Novel Software Technology, Nanjing University 
\\
$^2$ESAT, KU Leuven \quad $^3$Ant Group \quad $^4$Shanghai Artificial Intelligence Laboratory
}

\begin{document}

\makeatletter
\let\@oldmaketitle\@maketitle
\renewcommand{\@maketitle}{\@oldmaketitle
 \captionsetup{type=figure}
 \centering
  \includegraphics[width=0.9\linewidth]
    {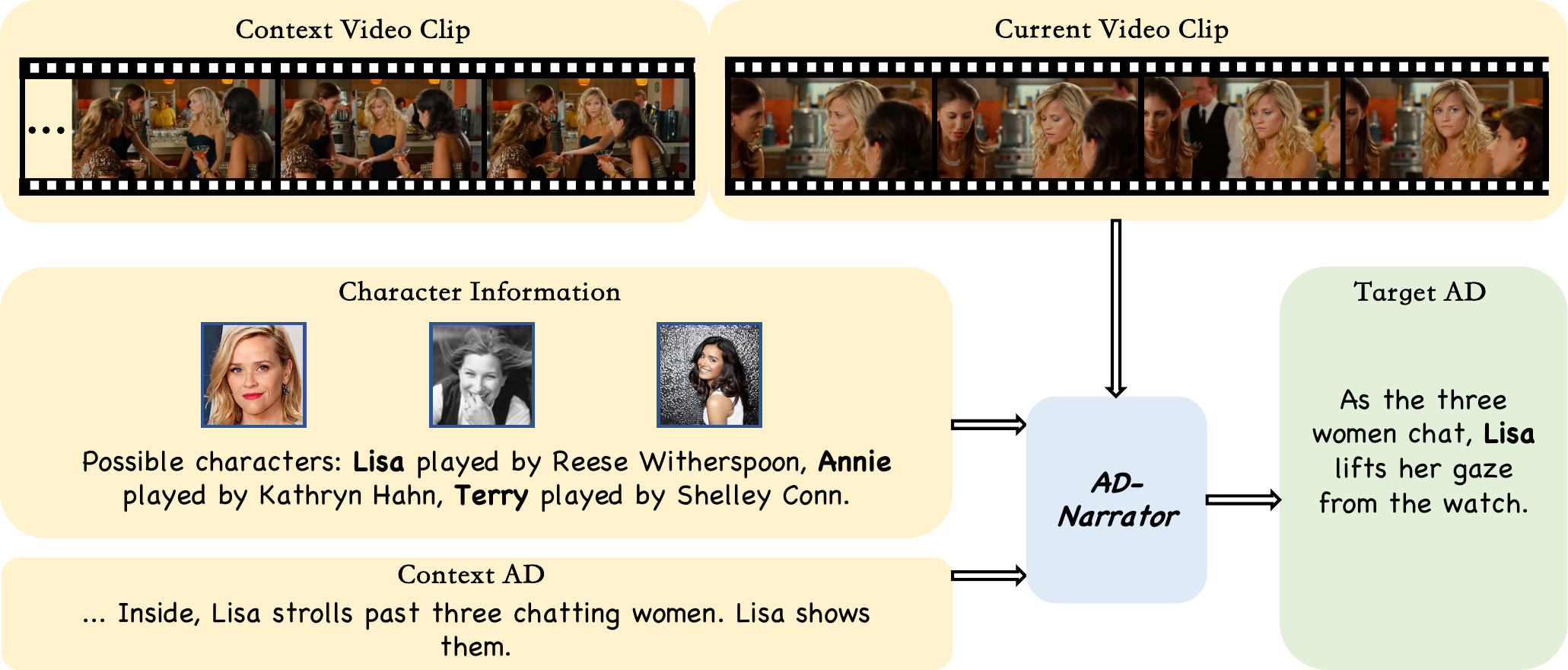}
    \bigskip
    \captionof{figure}{Taking video clip, text, character bank and context information as the inputs, the narrator generates corresponding audio description (AD) for video comprehension. Rather than describe all characters appearing in the video, the narrator should focus on characters that truly contribute to the storyline.
    }
    \vspace{5mm}
    \label{fig:teaser}   
    }
    
\maketitle

\let\thefootnote\relax\footnotetext{\noindent$^\dagger$Corresponding author.}

\begin{abstract}

The Audio Description (AD) task aims to generate descriptions of visual elements for visually impaired individuals to help them access long-form video content, like movies. With video feature, text, character bank and context information as inputs, the generated ADs are able to correspond to the characters by name and provide reasonable, contextual descriptions to help audience understand the storyline of movie. To achieve this goal, we propose to leverage pre-trained foundation models through a simple and unified framework to generate ADs with interleaved multimodal sequence as input, termed as \ours{}. To enhance the alignment of features across various modalities with finer granularity, we introduce a simple and lightweight module that maps video features into the textual feature space. Moreover, we also propose a character-refinement module to provide more precise information by identifying the main characters who play more significant roles in the video context. With these unique designs, we further incorporate contextual information and a contrastive loss into our architecture to generate smoother and more contextually appropriate ADs. Experiments on multiple AD datasets show that \ours{} performs well on AD generation, which demonstrates the effectiveness of our approach. 
Our code is available at:
\textcolor{magenta}{\href{https://github.com/ant-research/UniAD}{https://github.com/ant-research/UniAD}}.

\end{abstract}

\section{Introduction}
\label{sec:intro}

Audio Description (AD)~\cite{ad0, ad1, ad2, AutoAD0} provides descriptive narration of visual content in videos.
Unlike subtitle or transcription, AD focuses more on describing the scene, characters, actions and storyline of the input video. As a rich visual description, AD can effectively supplement the dialogue and provide viewers with a comprehensive description of the video content, which not only helps the visual impairments better engage with video content~\cite{ad0}, but also benefits the sighted individuals in their media comprehension activities~\cite{ad3, ad4}, such as language learning for kids and sight-free video consuming while driving. 
Despite that AD is important for video comprehension, particularly for those professionally produced media contents (movies, TV series \etc), currently most videos do not have corresponding AD~\cite{HCI}, mainly due to the considerable costs of manual annotation and differences in understanding between annotators~\cite{autoad2}. Therefore, studying how to generate ADs automatically is quite meaningful and necessary.

With the advances in computer vision and natural language processing, nowadays the research community is paying growing attention to generating ADs automatically, which requires a model to understand multi-modal information and perform contextual reasoning over video storyline~\cite{autoad, MM-Narrator}. Compared with the conventional video captioning task~\cite{caption1,caption2,caption3,caption4,caption5}, Audio Description (AD) is not only a scene description of the video clip, but also a narration that includes characters' names and actions to generate a coherent plot description, as shown in \cref{fig:teaser}. This brings two characteristics of the AD generating task: (i) Multiple modality inputs. Video clip, text, character portraits and names are provided for AD generating. (ii) Rich contextual information. Context video and past AD can be applied to assist the current AD generation.

Previous methods~\cite{autoad, autoad2} introduce learnable adapters into GPT-2 to generate ADs. However, the amount of parameters in these adapters will rapidly increase with larger LLM~\cite{autoad2}, thus not conducive to scaling up. Instead, AutoAD-III~\cite{Autoad3} applied Q-former architecture to bridge the visual space with the language space. Training-free methods like MM-Narrator~\cite{MM-Narrator} and AutoAD-Zero~\cite{AutoAD0} directly prompt GPT-4 or LLaMA3~\cite{LLaMA3} with specialized expert tools for AD generation, which suffers from complex prompt engineering and hallucination problem. In this work, we propose \ours{}, a simple architecture that takes interleaved multimodal sequence as input to leverage completely open-source LLMs~\cite{gpt2, llama} for AD generation by aligning various modality inputs to a unified semantic space. Formulating data as interleaved multimodal sequence makes it convenient to integrate various modality inputs and add contextual information for AD generation. Besides, interleaved multimodal sequence converts visual elements into multiple embedding tokens while maintaining the relative order between data, which ensures embeddings and tokens with the same semantic are naturally close to each other so that more fine-grained feature alignment can be learned spontaneously.

Given that the development of storyline is always character-centered~\cite{movie_char, char}, it is necessary for ADs to include character names to describe their expressions, actions and status. 
Previous method~\cite{autoad2} tried to identify all characters appearing in the given video as character information, without considering who are the main story drivers that should be included in AD. For example, in~\cref{fig:teaser}, though there are multiple characters in the scene, the expression change of \textit{Lisa} is the main content thus only her name is involved in the target AD. With such an observation, we in this paper design a character-refinement module to figure out the AD-related characters.
After training, this module can be applied to any videos to recognize main characters who contribute to the storyline and provide more accurate character information.

We further investigate the contextual information on our framework by combining past visual contents and ADs into the interleaved multimodal sequence, rather than only concatenating past ADs like previous works~\cite{autoad, autoad2, MM-Narrator}. We find that when the input video is similar to the past video clip, the model tends to generate identical ADs. To address this, we introduce a contrastive loss as an auxiliary to avoid repetition and encourage diversity in AD generation.

To sum up, we develop an AD narration system called \ours{} which achieves finer-grained feature alignment and supports extension to larger LLMs by formulating multiple inputs as interleaved multimodal sequence. To produce more accurate, coherent audio descriptions, we introduce a character-refinement module and incorporate contextual information along with a contrastive loss. Our \ours{} outperforms previous methods on multiple AD datasets.

\section{Related Work}\label{sec:related}

\subsection{Audio Descriptions Generation}
Audio Description (AD) describes the key visual elements in videos to form coherent storyline narration. With the development of media technology, captioning for videos has emerged as a growing area in the computer vision research community~\cite{videocaption1, videocaption2, caption1}. Nonetheless, the production of Audio Description (AD) for video content remains a relatively untapped area of research. Initial works designed specialized authoring tools~\cite{LiveDescribe} and evaluation mechanisms~\cite{CrossA11y, Viscene} to collect manually annotated ADs. Several annotation platforms like LiveDescribe~\cite{LiveDescribe}, Rescribe~\cite{Rescribe} also emerged to facilitate AD generation. Recently, some works have studied how to generate AD at scale automatically with deep learning models. AutoAD-\uppercase\expandafter{\romannumeral1}~\cite{autoad} introduced the task of AD generation for movies and addressed it by prompting GPT-2 with learnable visual prompt vectors. ~\cite{autoad2} later incorporated an external character bank to provide character information for more accurate AD generation. Researchers further applied Q-former architecture to bridge the visual and language space~\cite{Autoad3} for this task. Training free methods~\cite{MM-Narrator, AutoAD0} proposed designs which extract information from inputs with multimodal experts and queries GPT-4~\cite{GPT4} or VideoLLaMA2~\cite{videollama2} in a few-shot manner. However, these methods suffer from drawbacks like poor scalability, complex prompt engineering and weak modality alignment, which we in this paper address with the interleaved multimodal sequence design and larger LLM.

\subsection{Interleaved Sequence for Multimodal Learning}
Traditional vision language datasets for multimodal learning are mainly composed of image-text pairs collected from Internet~\cite{CC3M, CLIP}. Text contents in these datasets are mostly short, less descriptive and independent, resulting in relatively poor text embeddings. Recent works like Flamingo~\cite{Flamingo}, BLIP-2~\cite{blip2} and CM3~\cite{CM3} presented to conduct learning on the entire multimodal webpages, formulating interleaved images, videos and text as cohesive sequences. Such sequences offer long-form visual-text pairs for modeling and naturally retain the semantic correlation between different modality information, boosting the development of multi-modal learning. Other works ~\cite{Emu, DreamLLM, COSMO} conducting pre-training on large amounts of interleaved multi-modal data further demonstrate the importance and effectiveness of this way. Inspired by these works, we cast inputs for audio description as interleaved multimodal sequences, leveraging the semantic relevance to achieve finer-grained modality alignment.

\subsection{LLMs for Video Understanding} 
The recent surge in large language models (LLMs)~\cite{llms, llama, gpt2, gpt3, GPT4} has inspired the study of video perception and understanding with LLMs. Models like ChatCaptioner~\cite{VideoChatCaptioner}, VideoChat~\cite{VideoChat} and MM-Vid~\cite{MM-VID} integrate visual experts with LLMs to construct multimodal perception systems for video representation, long-term video comprehension and dialog-centric interaction, \etc. All these works can be divided into three categories:
(i) Prompt tuning~\cite{ClipCap, VideoChatCaptioner} is a lightweight approach to transfer LLMs to downstream tasks with learnable prompt vectors. (ii) Adapter-based methods~\cite{VideoChat} typically insert additional trainable parameters into the LLM at different positions to achieve deep modality alignment, but the amount of introduced parameters increases with the size of LLM, making adapter-based methods difficult to scale up. (iii) Querying LLMs in a training free manner~\cite{MM-VID}, which employs visual experts to transform video into text, thereby guiding LLMs in reasoning on specific tasks. Our \ours{} follows the visual-conditioned prompt tuning manner to extend our approach to larger LLMs with its memory-friendly characteristic to generate better ADs.

\begin{figure*}[t]
  \centering
  \includegraphics[width=1.0\textwidth]{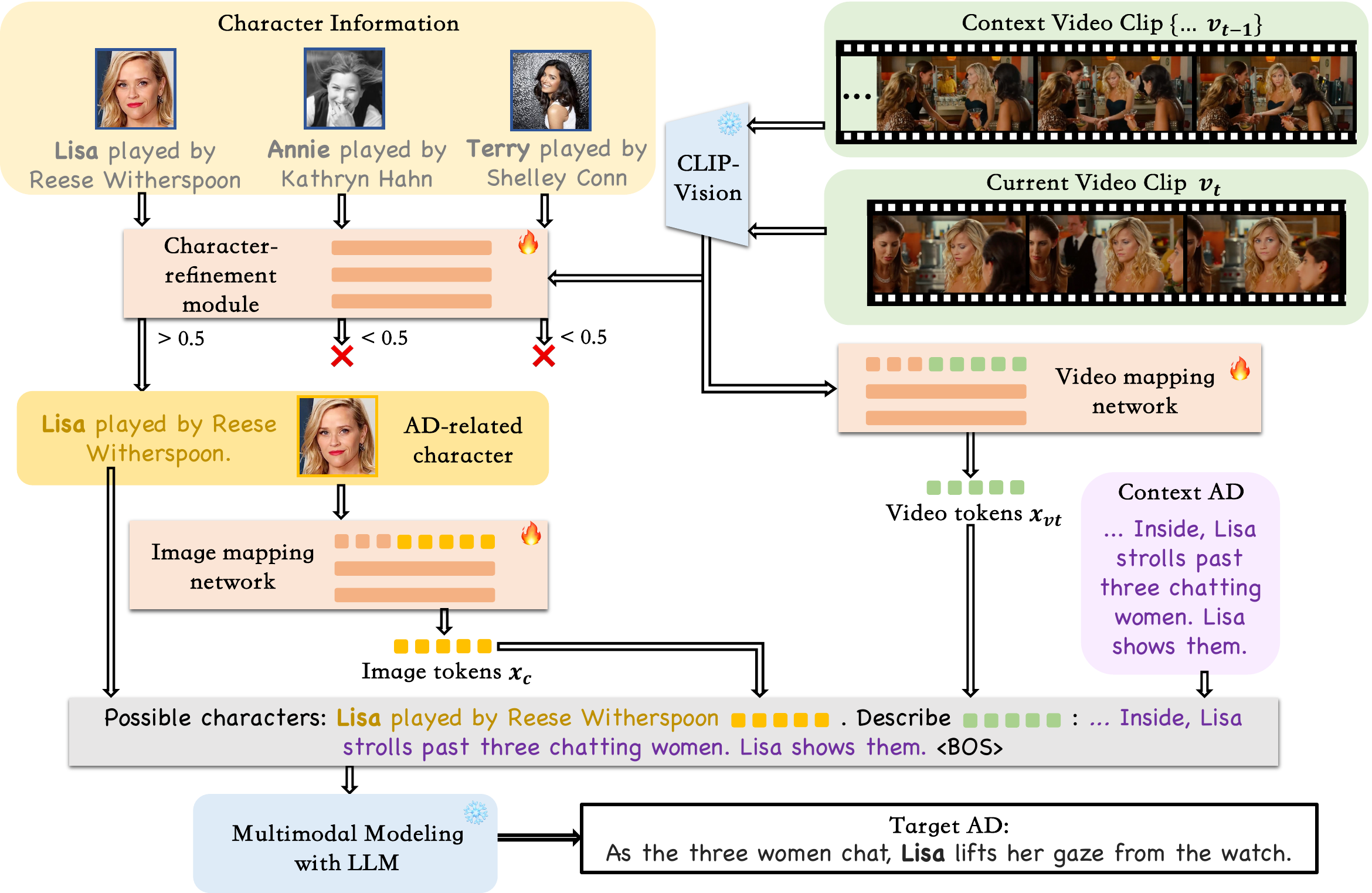}
  \caption{Overall architecture of our proposed \ours{}. Our model first filters the input character information to retain the AD-related characters. Then all visual contents are mapped into the unified semantic space to form the interleaved multimodal sequence with text and contextual information. Afterwards, we prompt a frozen LLM with this sequence to generate the corresponding AD.}
  \label{fig:overview}
  \vspace{-10pt}
\end{figure*}

\section{Methodology}\label{sec:method}

The audio description (AD) task is challenging mainly due the requirement that generated results should include characters' names to depict their expressions, actions to advance the plot and should be contextually coherent within the storyline. To meet this, we present \ours{} to formulate various inputs for current video clip along with contextual information as interleaved multimodal sequence and combines a character-refinement module. In this section, we will give a detailed description of our approach. First, we provide an overview of \ours{} in \cref{subsec::overview}. Then we present how to prompt a LLM for AD generation in \cref{subsec::interleaved}. Next, we show the design and training details of our proposed character-refinement module in \cref{subsec::character}. Afterwards, we describe the usage of contextual information for AD generation in \cref{subsec::contextual information}. Finally, in \cref{subsec::training scheme} we further introduce the learning object of our framework.

\subsection{Method Overview}
\label{subsec::overview}

The overall framework of our \ours{} is illustrated in \cref{fig:overview}, which
contains two key stages: visual modality alignment and multimodal prompt generation. Given a video clip and its corresponding character, contextual information, our model first filters the input character information to retain the AD-related
individuals. Then we map all visual contents into embeddings, which will be combined with text tokens to create prompt for LLM. Contextual information can also be involved in this sequence. Finally, this interleaved multimodal prompt will be fed to a frozen LLM to generate audio description. Contrastive loss can be applied to avoid generating duplicate ADs.

\subsection{Uni-AD Pipeline}
\label{subsec::interleaved}

\begin{figure}[t]
  \centering
  \includegraphics[width=\linewidth]{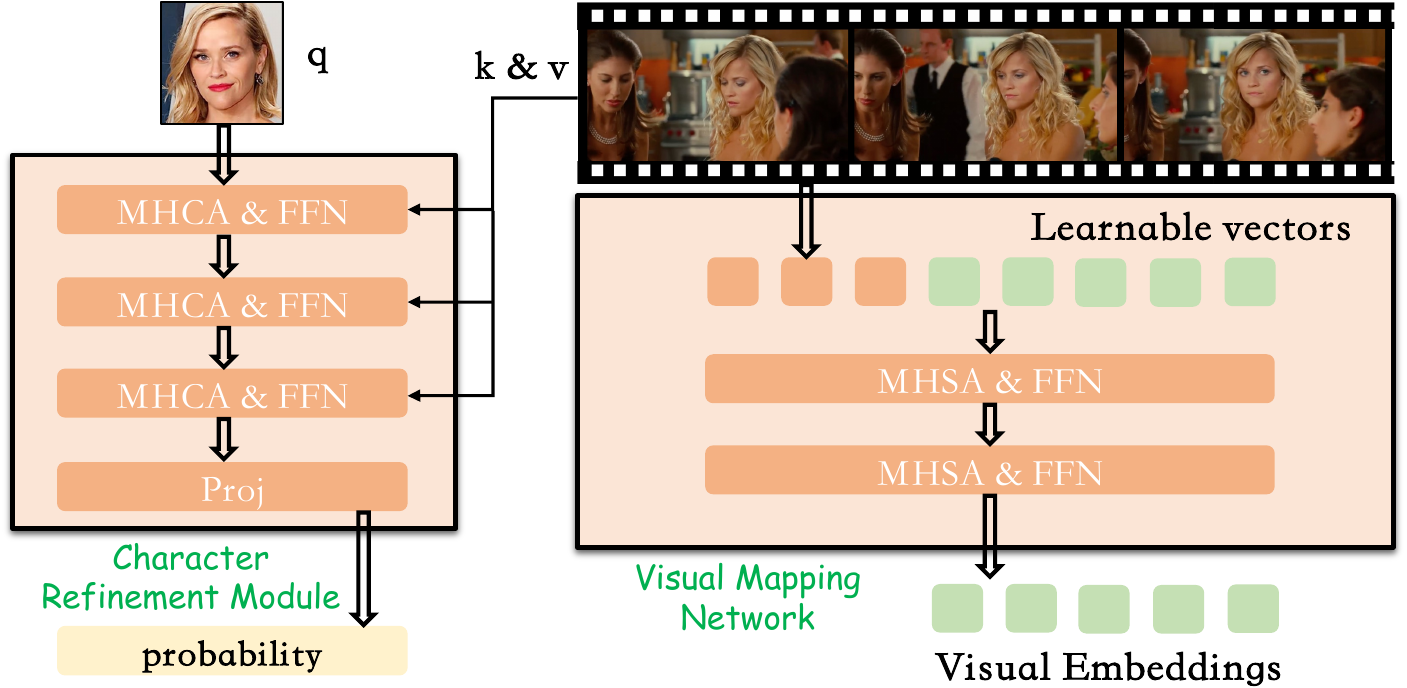} 
   \caption{Structure of Visual Mapping Network and Character-Refinement Module.}
   \label{fig:structure}
   \vspace{-10pt}
\end{figure}

\noindent \textbf{Visual Mapping Network.}
The visual input for AD generation includes two components: (1) video clip consisting of $N$ frames with timestamp $t$, denoted as $v_t = \{\mathcal{I}_1,\mathcal{I}_2,...,\mathcal{I}_N\}$; (2) characters' portrait images, denoted as $\{\mathcal{A}_1,\mathcal{A}_2,...,\mathcal{A}_C\}$, where $C$ is the number of related characters in video clip $v_t$. To produce corresponding AD for $v_t$, we need to transform visual elements into embeddings with visual mapping network to achieve cross-modal alignment. Inspired by ClipCap~\cite{ClipCap} and AutoAD-\uppercase\expandafter{\romannumeral1}~\cite{autoad}, we here apply a multi-layer transformer encoder with a fixed number of learnable vectors as our mapping network (shown in \cref{fig:structure}) based on the following findings: First, video frames are able to interact with each other to model the temporal relation via the attention mechanism, which is essential given that effective interaction between visual tokens is hard to achieve in a frozen LLM. Second, the introduction of learnable vectors allows us to control the length of the visual representation and using more embeddings to preserve visual details. Finally, such a structure requires no addition of adapters inserted into the LLM for modality alignment, making it easier to scale to larger LLMs compared to methods like AutoAD-\uppercase\expandafter{\romannumeral2}~\cite{autoad}. 

Specifically, we first extract visual features of the input video clip and character images with the pre-trained CLIP~\cite{CLIP} visual encoder:
\begin{equation}
    \begin{split}
        z_{v_t} &= f_{CLIP}(v_t), \\
        \{z_1, z_2, ..., z_C\} &= f_{CLIP}(\{\mathcal{A}_1,\mathcal{A}_2,...,\mathcal{A}_C\}).
    \end{split}
\end{equation}

In order to reduce the impact of difference between an actor's portrait and appearance in films, we follow ~\cite{autoad2} to adopt exemplar feature as character image information. The exemplar feature is obtained by averaging features of 5 frames that are most similar to the actor's portrait within the same movie. CLIP feature of the $i_{th}$ character
$z_i$ is used as signature to compute similarity with movie CLIP features. Afterwards, we convert the visual input into embeddings with our visual mapping network $\mathcal{M}_v$:
\begin{equation}
    \begin{split}
         x_{v_t} &= \mathcal{M}_v(Proj(z_{v_t})), \\
         \{x_1, x_2, ..., x_C\} &=
        \mathcal{M}_v(Proj(e_1),...,Proj(e_C)),
    \end{split}
\end{equation}
where $e_i$ denotes the $i_{th}$ character's exemplar feature and $Proj$ represents a Linear Layer that transforms the channel number of visual features to match the LLM. Note that exemplar features are mapped via $\mathcal{M}_v$ separately, meaning there is no need for interaction between characters here.

\noindent \textbf{Formulating Interleaved Multimodal Prompt.}
\label{prompt}
With visual embeddings $x_{v_t}$ and $\{x_1, x_2, ..., x_C\}$, we now combine them with the text query to get our interleaved multimodal prompt. We apply the prompting template in ~\cite{autoad2} 
to query the frozen LLM, and our prompt is formulated as:
\begin{center}
\noindent $\langle$ Possible characters: $char_1$ played by $actor_1$ $x_1$, $char_2$ played by $actor_2$ $x_2$, ... Describe $x_{v_t}$: $\rangle$,
\end{center}
\noindent where $char_i$, $actor_i$, $x_i$, $x_{v_t}$ denote the $i_{th}$ character's name, real actor name, image feature and video embeddings, respectively. Our main thought here is to represent visual content with multiple tokens and preserve the positional relationships between different modalities of information, hoping to achieve a finer-grained alignment.

\subsection{Character-Refinement Module}
\label{subsec::character}

Character information provides names, portraits of active characters to help model generate person-centric descriptions, making it important to be incorporated into AD generation. Previous work achieved this by introducing an external character bank~\cite{autoad2}. With an aim to recognize all \textit{active} characters who appear in the given video clip, researchers trained a character recognition module on the annotated MovieNet~\cite{MovieNet} dataset to predict the active characters given their exemplars and the movie clip. Then the output of this module is used to build the character bank to provide character information for AD generation.

Introducing the external character bank significantly improves the quality of AD generation. However, this approach overlooks the difference between \textit{active} characters and \textit{AD-related} characters. That is, a character who appears in the current video clip may not necessarily be mentioned in the corresponding AD, especially in scenes with multiple characters. ADs usually prioritize the main characters who drive the story forward and do not mention secondary characters to avoid overburdening audiences with too much information. In such cases, taking all characters appearing in the video as character information will confuse the AD narrator and generate descriptions not align with the development of the storyline.

Given observations mentioned above, we adapt our goal to identify the \textit{AD-related} characters in the video clip based on their behavior and mannerisms. Since these characters are likely to appear in the video clip, we design a character-refinement module to perform further identification based on the external character bank provided by AutoAD-\uppercase\expandafter{\romannumeral2}~\cite{autoad2}. Our character-refinement module consists of 3 Multi-Head Cross-Attention (MHCA) and FeedFoward Network (FFN) layers (shown in \cref{fig:structure}), which takes characters' exemplar features in external character bank as query, video features as key$\&$value and outputs the probability for each character on whether they are AD-related.
A projection layer is added to transform the output feature to a probability value. We train this module with a binary classification loss. Training labels are obtained from the AD annotations by retrieving all names that appear in both the movie's cast list and the annotation~\cite{autoad2}. After trained, we apply this module to the test movies and treat characters whose probability exceeds 0.5 as AD-related characters.

\subsection{Contextual Information Modeling}
\label{subsec::contextual information}

We in this section show how to use contextual information to generate more coherent ADs with our model. With the design of interleaved multimodal sequence, we can easily incorporate past context ADs and video clips into our prompt for LLM.

Context ADs contain descriptions of preceding story plot that leads up to the current scene, thus can help the model better follow the storyline for narration. We here utilize context AD by directly concatenating our prompt in \cref{prompt} with the past $K$ ADs $\{\mathcal{T}_{t-K}, \mathcal{T}_{t-K}, ..., \mathcal{T}_{t-1}\}$. In this way, we provide more text conditional information for AD generation. To separate context ADs from current AD, we add a $BOS$ token in our prompt to start AD generation. 

Although context ADs can provide the most accurate description of the preceding story plot, the ADs we generate during inference will inevitably differ from the ground truth, leading to an inaccurate guidance. Therefore, we consider to introduce past videos into our \ours{} as contextual information. Specifically, we take frame features from the past $K$ video clips $\{z_{v_{t-K}}, z_{v_{t-K+1}}, ..., z_{v_{t-1}}\}$ and concatenate them with the current video clip in temporal order. Then these concatenated visual features will be fed to the visual mapping network $\mathcal{M}_v$ for interaction, resulting in the contextual video representation $x_{t\_context}$:
\begin{equation}
    x_{t\_context} = \mathcal{M}_v(Proj(z_{v_{t-K}}; z_{v_{t-K+1}}; ...; z_{v_t})),
\end{equation}
where $[\cdot;\cdot]$ denotes the concatenation operation. Then we can use $x_{t\_context}$ instead of $x_{v_t}$ as video embedding to get our video-context prompt. \cref{fig:overview} shows our prompt with both visual and text contextual information.

In practice, we find that when the input video does not vary much from the past video clip, LLM tends to generate very similar, or even identical ADs. This is unreasonable for the AD generation task, since current AD should carry on the narration from previous content rather than repeating. To address this, we add a contrastive loss to our training process:
\begin{equation}
\label{eq:cl}
    \begin{split}
        s = \frac{\sum_{n}logP_{\Theta}(a_n| \mathrm{prompt}; a_{<n})}{||\mathcal{T}||},\\
        \mathcal{L}_{ct} = max(0, s_{last} - s_{current}),
    \end{split}
\end{equation}
where $\mathcal{T}$ denotes an AD, $a_n$ denotes the $n_{th}$ token in $\mathcal{T}$, $a_{<n}$ denotes tokens preceding $a_n$ in $\mathcal{T}$; prompt denotes our interleaved multimodal sequence fed to LLM; $\Theta$ denotes learnable parameters in our visual mapping network; $s$ represents the average likelihood score of the generated AD. Our contrastive loss $\mathcal{L}_{ct}$ is calculated as \cref{eq:cl}. Thus our model is constrained to ensure that the score of generating last AD ($s_{last}$) is always lower than score of the current ground-truth AD ($s_{current}$). In this way, we encourage the model to generate more accurate and non-repetitive ADs.

\subsection{Objective Function}
\label{subsec::training scheme}

Overall, given a video clip with timestamp $t$, our goal is to query a frozen LLM for AD generating with our visual-conditioned prompt. The supervision we apply is the commonly used auto-regressive loss function:

\begin{equation}
    \mathcal{L}_{auto} = -\sum_{n} \log P_{\Theta}(a_n|\mathrm{prompt}; a_{<n}),
\end{equation}
where $a_n$ denotes the $n_{th}$ token in the target AD and $prompt$ denotes our interleaved multimodal prompt.

As mentioned in \cref{subsec::contextual information}, we further introduce a contrastive loss to avoid repetitive AD generation. In this case, our complete loss is: $\mathcal{L}_{\Theta} = \mathcal{L}_{auto} + \mathcal{L}_{ct}$. 

\section{Experiments}\label{sec:exp}

In this section, we evaluate \ours{} on multiple AD generation benchmarks and show the experiment results. We first introduce our implementation details in \cref{exp::Experiment Settings}. Then we compare our performance with state-of-the-art AD generation approaches in \cref{exp::Comparison}. We further conduct a detailed ablation study on the impact of character-refinement module and visual mapping network on our model in \cref{exp::ablation}. Next in \cref{exp::context}, we confirm the effectiveness of incorporating contextual information into \ours{} and show how different contextual information affects our model. Finally, we provide qualitative examples of our \ours{} in \cref{exp::example}.

\subsection{Implementation Details}
\label{exp::Experiment Settings}

\noindent \textbf{Dataset.}
We follow AutoAD-\uppercase\expandafter{\romannumeral1}~\cite{autoad} to conduct partial-data pre-training on the AudioVault-AD dataset~\cite{autoad}. We train our model on the MAD-v2-Named dataset and evaluate on MAD-eval-Named~\cite{MAD, autoad}. For evaluation on CMDAD~\cite{Autoad3} and TVAD~\cite{AutoAD0}, we train our model with the CMDAD~\cite{Autoad3} training set. We use both classic captioning metrics and newly proposed metrics for evaluation. The former metrics include ROUGE-L~\cite{rouge} (R-L) and CIDEr~\cite{cider} (C) to measure the quality of our generated ADs versus human-annotated ones. The latter metrics include R@k/N~\cite{autoad2}, CRITIC~\cite{Autoad3} and LLM-AD-eval~\cite{Autoad3}. More information about datasets and metrics is provided in the supplementary material.

\noindent \textbf{Training details.}
We train our GPT-based model with a batch size of 96 movie clips and the learning rate is $10^{-3}$, while our LLaMA-based model is trained with a batch size of 12 movie clips and the learning rate is $5.0 \times 10^{-5}$. We use the AdamW~\cite{adamw} optimizer to train our model for 10 epochs, with a cosine-decayed learning rate schedule and linear warm-up. All training is conducted on 8 A100 GPUs. For external character information, we use the prediction results from AutoAD-\uppercase\expandafter{\romannumeral2}~\cite{autoad2} as input for our character refine module on MAD-eval-Named dataset and prediction results from AutoAD-Zero~\cite{AutoAD0} as input for character refine module on CMDAD and TVAD. We apply the visual branch of VideoLLaMA~\cite{videollama} as visual mapping network on experiments of CMDAD and TVAD, as in ~\cite{Autoad3}.

\subsection{Comparison with state-of-the-art approaches}
\label{exp::Comparison}
\begin{table}[t]
\caption{Comparison with the state-of-the-art methods on MAD-eval-Named. The \textit{Context} column denotes whether contextual information is applied. The \textit{V-Feature} column indicates the type of visual expert used for extracting movie frame features.}
\label{tab::comparison}
\centering
\resizebox{1.0\linewidth}{!}{
\setlength{\tabcolsep}{2pt}
\vspace{-3pt}
\begin{tabular}{c|cc|ccc}
\toprule
\textbf{Methods}     & \textbf{Context}    & \textbf{LLM \& V-Feature} & \multicolumn{1}{c}{\textbf{RL$\uparrow$}}   & \multicolumn{1}{c}{\textbf{C$\uparrow$}}    & \multicolumn{1}{c}{\textbf{R@5/16$\uparrow$}} \\ \hline

ClipCap~\cite{ClipCap}     &    \xmark            & GPT-2 \& CLIP-B32    & \multicolumn{1}{c}{8.5} & \multicolumn{1}{c}{4.4} &  36.5      \\
AutoAD-\uppercase\expandafter{\romannumeral1}~\cite{autoad}      &    \xmark     & GPT-2 \& CLIP-B32    & 10.3 & 12.1 &  39.8      \\
AutoAD-\uppercase\expandafter{\romannumeral2}~\cite{autoad2}     &    \xmark     & GPT-2 \& CLIP-B32    & 13.1 & 19.2 &   51.3     \\
AutoAD-\uppercase\expandafter{\romannumeral3}~\cite{Autoad3}     &    \xmark     & LLaMA2 \& EVA-CLIP    & -   & 24.0     &    52.8     \\
AutoAD-Zero~\cite{AutoAD0}    &    \xmark     & LLaMA3 \& VideoLLaMA2    & -   &  22.4   &   -    \\
\ours{}(ours)       &    \xmark     & GPT-2 \& CLIP-B32    & 15.9 & 24.0 &  50.5      \\
\ours{}(ours)       &    \xmark     & GPT-2 \& CLIP-L14    & 16.4 & 25.7     &  51.5      \\
\ours{}(ours)       &    \xmark           & LLaMA2 \& CLIP-L14   & \textbf{16.8} &  \textbf{27.3}    &   \textbf{53.3}     \\ \hline
AutoAD-\uppercase\expandafter{\romannumeral1}~\cite{autoad}      &    \cmark      & GPT-2 \& CLIP-B32    &   11.9   & 14.3     &    42.1    \\
AutoAD-\uppercase\expandafter{\romannumeral2}~\cite{autoad2}     &    \cmark      & GPT-2 \& CLIP-B32    &   13.4   &  19.5    &   50.8     \\
MM-Narrator~\cite{MM-Narrator} &    \cmark               & GPT-4 \& CLIP-L14    & 13.4 &   13.9   &   49.0     \\
MM-Narrator~\cite{MM-Narrator} &    \cmark              & GPT-4V \& CLIP-L14    & 12.8 &   9.8   &   -     \\
\ours{}(ours)       &    \cmark          & LLaMA2 \& CLIP-L14   & \textbf{17.1} &  \textbf{28.2}    &   \textbf{54.2}     \\ \hline

\end{tabular}
}
\vspace{-8pt}
\end{table}

\begin{table}[t]
\vspace{6pt}
\caption{Comparison with the state-of-the-art methods on CMDAD and TVAD. The gray row shows the results of AutoAD-III pretrained on the 3.4M HowTo-AD dataset, which is not public.}

\label{tab::comparison2}
  \centering
  \resizebox{0.47\textwidth}{!}{
  \begin{tabular}{c|c|ccc}
  \toprule
    \textbf{Method} & \textbf{Dataset} & \textbf{CIDEr$\uparrow$} &  \textbf{CRITIC$\uparrow$} & \textbf{LLM-AD-eval$\uparrow$} \\
    \hline
    AutoAD-II~\cite{autoad2} & CMDAD & 13.5 &  8.2 & 2.08  \\
    AutoAD-III~\cite{Autoad3} & CMDAD & 21.7  & 25.2 & 2.85 \\
    \rowcolor{gray!40} AutoAD-III~\cite{Autoad3} & CMDAD & 25.0 & 32.7 & 2.92 \\
    AutoAD-Zero~\cite{AutoAD0} & CMDAD & 17.7 & \textbf{43.7} & 2.83 \\
    \ours{}(Ours) & CMDAD & \textbf{21.8}  &  41.9 & \textbf{2.92} \\ \hline
    AutoAD-III~\cite{Autoad3} & TVAD & 26.1  &  15.9 & 2.78  \\
    AutoAD-Zero~\cite{AutoAD0} & TVAD & 22.6 &  27.6 & \textbf{2.94} \\
    \ours{}(Ours) & TVAD & \textbf{26.6} &  \textbf{28.3} & 2.89 \\ \hline
  \end{tabular}
  }
  \vspace{-15pt}
\end{table}

Multiple models for AD generation are involved in our comparison. Descriptions for these methods are available in the supplementary material. \cref{tab::comparison} shows the evaluation results of our \ours{} and these state-of-the-art methods on the MAD-eval-Named benchmark. For fair comparison, we first evaluate \ours{} under the same setting~(GPT-2 as language decoder and CLIP ViT-B/32 as visual feature) with previous methods~\cite{ClipCap,autoad,autoad2}. Our \ours{} outperforms all previous methods by a notable margin, which demonstrates the effectiveness of our model. Afterward, we apply a stronger CLIP ViT-L/14 model as visual encoder to extract movie features, and the results of \ours{} show growth on all metrics. Utilizing the design of lightweight visual mapping network, we then extend our method to a more powerful LLM LLaMA2-7B, which further enhances the performance~(1.6 growth points on \textit{Cider}, 0.4 on \textit{Rouge-L} and 1.8 on \textit{Recall@5/16}).

Next, we conduct comparison with AD generation approaches under the contextual setting. \ours{} achieves the state-of-the-art performance~(17.1 on \textit{Rouge-L}, 28.2 on \textit{Cider} and 54.2 on \textit{Recall@5/16}) by incorporating context video features and contrastive learning. We further evaluate \ours{} on CMDAD and TVAD and obtain competitive results, shown in \cref{tab::comparison2}. These outstanding performances show the effectiveness and flexibility of our model.

\subsection{Ablation Study}
\label{exp::ablation}

\begin{table}[t]
\caption{Study the effectiveness of character-refinement module without pre-training or contextual information. \textit{Char.?} shows whether this module is applied. * means our re-implemention.}
\label{tab::char}
\centering
\resizebox{0.85\linewidth}{!}{
\setlength{\tabcolsep}{6pt}
    \begin{tabular}{c|c|ccc}
    \toprule
    \textbf{Methods}   & \textbf{Char.?} & \multicolumn{1}{c}{\textbf{RL$\uparrow$}}   & \multicolumn{1}{c}{\textbf{C$\uparrow$}}    & \multicolumn{1}{c}{\textbf{R@5/16$\uparrow$}} \\ \hline
    \multirow{3}{*}{\shortstack{AutoAD-\uppercase\expandafter{\romannumeral2}
    \\(GPT-2-B32)~\cite{autoad2}}}  & \xmark      & \,\,\,14.7$^*$    & \,\,\,19.0$^*$  & \,\,\,46.0$^*$   \\
    &  \cmark       & 15.1    & 21.0  & 47.8   \\
 &   \cellcolor{mygray}\textit{gt}    & \cellcolor{mygray}19.4    & \cellcolor{mygray}33.8  & \cellcolor{mygray}68.0  \\ \hline
 \multirow{3}{*}{\shortstack{\ours{} \\ (GPT-2-B32)}}   &  \xmark      & 14.0    & 18.5  &    45.7    \\
&   \cmark       & 15.7    & 23.7  & 49.4   \\
 &   \cellcolor{mygray}\textit{gt}    & \cellcolor{mygray}19.7    & \cellcolor{mygray}36.1  & \cellcolor{mygray}69.2   \\ \hline
\multirow{3}{*}{\shortstack{\ours{} \\ (GPT-2-L14)}} & \xmark  &    14.0   &   19.7  &  47.2   \\
& \cmark       & 16.0    & 25.5  & 50.3   \\
 & \cellcolor{mygray}\textit{gt}    & \cellcolor{mygray}20.3    & \cellcolor{mygray}37.8  &   \cellcolor{mygray}70.4     \\ \hline
\multirow{3}{*}{\shortstack{\ours{} \\ (LLaMA2-L14)}} & \xmark  & 15.3    & 22.6  &    \textbf{53.1}    \\
&  \cmark       & \textbf{16.5}    & \textbf{25.9}  & 52.5   \\
 &  \cellcolor{mygray}\textit{gt}   & \cellcolor{mygray}21.2    & \cellcolor{mygray}40.0  & \cellcolor{mygray}72.0   \\ \hline
    \end{tabular}
    }
    \vspace{-10pt}
\end{table}

\noindent \textbf{Study on character-refinement module.}
\label{exp::character}
By recognizing the main characters contributed to the storyline, the character-refinement module provides more precise character information for AD generation. Evaluation results in \cref{tab::char} show the general performance improvements character-refinement module brings under different settings, confirming the universality of this module.

We further use characters involved in AD annotations as character information to conduct experiments~(\textit{gt} rows in \cref{tab::char}). The gap between character-refinement results and \textit{gt} results also show that the current character-refinement performance~(0.41 on \textit{Precision}, 0.77 on \textit{Recall}, evaluated on MAD-eval-Named dataset) is far from sufficient and there is still significant room for improvement.

\noindent \textbf{Study on Visual Mapping Network.}
\label{exp::visual}
To verify whether our \ours{} can achieve finer-grained feature alignment, we in this section study how the visual mapping network affects AD generation. Our main comparison target is the Flamingo-style method AutoAD-\uppercase\expandafter{\romannumeral2}. Specifically, we adjust the number of latent vectors output by the visual mapping network, which also corresponds to the number of visual tokens fed to the LLM, to observe the impact on AD generation. The evaluation results are shown in \cref{tab::visual}. It can be seen that as the number of latent vectors increases, the performance of \ours{} improves under both settings (GPT-CLIP-B32 and LLaMA-CLIP-L14), while the result of AutoAD-\uppercase\expandafter{\romannumeral2} method remains basically unchanged. This confirms our hypothesis that AutoAD-\uppercase\expandafter{\romannumeral2}, which achieves cross-modal alignment by concatenating characters' portraits with video frames followed by perceiver resampler and gated cross-attention modules, tends to extract global feature of the visual contents. In contrast, our \ours{} is capable of retaining more visual details by increasing the number of latent vectors, thereby achieving finer-grained feature alignment and better AD generation results.

\begin{table}[t]
\caption{Study the impact of visual mapping network on AD generation without pre-training or Contextual information. Character-refinement module is applied by all methods. \textit{\#Latent} denotes the number of learnable vectors in the visual mapping network.}\label{tab::visual}
\centering
\resizebox{0.8\linewidth}{!}{
    \setlength{\tabcolsep}{6pt}
    \begin{tabular}{c|c|ccc}
    \toprule
    \textbf{Methods}  & \textbf{\#Latent}  & \multicolumn{1}{c}{\textbf{RL$\uparrow$}}   & \multicolumn{1}{c}{\textbf{C$\uparrow$}}    & \multicolumn{1}{c}{\textbf{R@5/16$\uparrow$}} \\ \hline
    \multirow{4}{*}{\shortstack{AutoAD-\uppercase\expandafter{\romannumeral2}~\cite{autoad2} \\ (GPT-2-B32)}} & 1         & 15.1    & 20.5  & 47.5   \\
    & 5         & 15.1    & 20.8  & 48.0   \\
    & 10        & 15.1    & 21.0  & 47.8   \\
    & 30        & 15.1    & 20.1  & 47.4   \\
    \hline
    \multirow{5}{*}{\shortstack{\ours{} \\ (GPT-2-B32)}}    & 1   & 13.7    & 19.0  & 45.4   \\
    & 5           & 15.3    & 22.8  & 47.9   \\
    & 10          & 15.4    & 22.4  & 49.0   \\
    & 30          & 15.7    & 23.7  & 49.4   \\
    \hline
    \multirow{5}{*}{\shortstack{\ours{} \\ (LLaMA-L14)}} & 1     & 16.2    & 24.9  & 51.5   \\
    & 5           & 16.3    & 24.9  & 52.0   \\
    & 10          & 16.4    & 25.0  & 52.0   \\
    & 30          & \textbf{16.5}    & \textbf{25.9}  & 52.5   \\
    & 60          & 16.3    & 25.3  & \textbf{53.7}   \\ \hline
    \end{tabular}
}
\vspace{-5pt}
\end{table}

\begin{table}[t]
\caption{Study on contextual information. We conduct experiments with our AudioVault pre-trained LLaMA-CLIP-L14 model.}\label{tab:context1}
\centering
\resizebox{0.75\linewidth}{!}{
    \setlength{\tabcolsep}{6pt}
    \begin{tabular}{cc|ccc}
    \toprule
\textbf{Context-V}   & \textbf{C-Loss} & \multicolumn{1}{c}{\textbf{RL$\uparrow$}}   & \multicolumn{1}{c}{\textbf{C$\uparrow$}}    & \multicolumn{1}{c}{\textbf{R@5/16$\uparrow$}} \\ \hline
0  & \xmark      &  16.8   & 27.3  &  53.3  \\
1  &  \xmark     &  16.8   & 27.5  &  54.7  \\
3 &   \xmark     &  17.0   & 27.4  &  54.2  \\ \hline
0 &   \cmark     &  16.9   & 27.3  &  53.7  \\
1 &   \cmark     &  \textbf{17.1}   & \textbf{28.2}  &  54.2  \\
3 &   \cmark     &  16.9   & 27.3  &  \textbf{54.9}  \\ \hline
\end{tabular}
}
\vspace{-12.5pt}
\end{table}

\subsection{Integrating Contextual Information}
\label{exp::context}

\begin{figure*}[h]
  \centering
  \includegraphics[width=\linewidth]{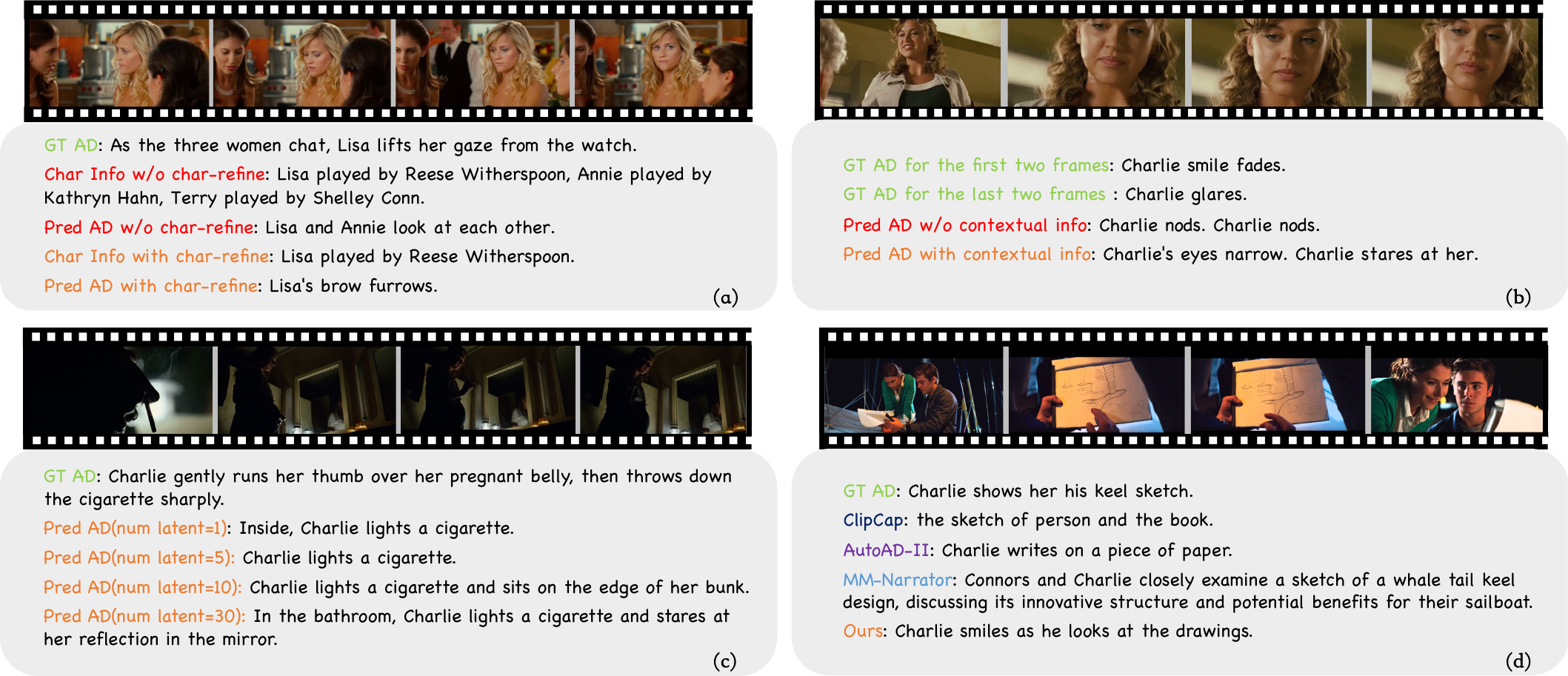} 

   \caption{Qualitative analysis on character-refinement module, contextual information, number of learnable vectors and comparison with other approaches. Movies are selected from (a): How Do You Know(2010), (bc): Legion(2010), (d): Charlie St. Cloud (2010).}
  \label{fig:examples}
\end{figure*}

For contextual information, we first study the effectiveness of integrating context video and contrastive loss into AD generation. Results in \cref{tab:context1} suggest that adding context video clips and contrastive loss can both enhance model's performance. Among these results, incorporating the most recent context video along with contrastive loss is the best, since current AD is most relevant to its preceding content.

Then we conduct experiments to introduce past ADs as additional contextual information under \textit{recurrent} setting (using predicted past ADs) and \textit{oracle} setting (using ground-truth past ADs). As shown in \cref{tab:context2}, the \textit{recurrent} setting leads to a decrease in model performance.
We attribute this to the discrepancy between the predicted ADs and the ground truth, which provides inaccurate information for AD generation during inference, thus further enlarges the gap between training and testing. For \textit{oracle} setting, the performance of our model improves with more context ADs, indicating that context AD is the most significant influencing factor for AD generation.

\begin{table}[t]
\vspace{1pt}
\caption{Study the effectiveness of context AD. \textit{Context-AD} denotes the number of past context ADs. \textit{V$\&$C-Loss} indicates whether past videos and contrastive loss are applied.}
\vspace{-2pt}
\label{tab:context2}
\centering
\resizebox{0.85\linewidth}{!}{
    \setlength{\tabcolsep}{6pt}
    \begin{tabular}{cc|ccc}
    \toprule
\textbf{Context-AD}   & \textbf{V\&C-Loss} & \multicolumn{1}{c}{\textbf{RL$\uparrow$}}   & \multicolumn{1}{c}{\textbf{C$\uparrow$}}    & \multicolumn{1}{c}{\textbf{R@5/16$\uparrow$}} \\ \hline
\multirow{2}{*}{1(recurrent)}
& \xmark      &  16.4  &  25.8 & 51.9   \\
& \cmark      &  16.4   & 25.8  & 52.3   \\ \hline
\multirow{2}{*}{3(recurrent)}
& \xmark      &  15.6  & 23.1  & 50.9    \\
& \cmark      &  16.1   & 24.0  & 52.2   \\ \hline
\multirow{2}{*}{1(oracle)}
& \xmark      &\cellcolor{mygray}17.7     &\cellcolor{mygray}31.3   &\cellcolor{mygray}55.3    \\
& \cmark      &\cellcolor{mygray}17.9      &\cellcolor{mygray}31.7   &\cellcolor{mygray}56.5    \\ \hline
\multirow{2}{*}{3(oracle)}
& \xmark      &\cellcolor{mygray}18.6     &\cellcolor{mygray}34.9   &\cellcolor{mygray}55.9    \\
& \cmark      &\cellcolor{mygray}18.6      &\cellcolor{mygray}34.8   &\cellcolor{mygray}55.6    \\ \hline
\end{tabular}
}
\end{table}

\subsection{Qualitative Results}
\label{exp::example}
We show our qualitative AD generation results on the MAD-eval dataset in \cref{fig:examples}. Specifically, we conduct analysis on character-refinement module, contextual information and the number of learnable vectors in visual mapping network. Results show that: (1) The character-refinement module can recognize the AD-related people and provide more precise character information for AD generation. For example in (a), AD-narrator without this module mistakes the main character as \textit{Lisa} and \textit{Annie}, thus generates AD deviated from the ground truth. (2) Incorporating contextual information and the contrastive loss can effectively avoid repeated AD generation and get more coherent results. In sample (b) where the contents of two consecutive movie clips vary little, model without contextual information just generates two identical ADs, while descriptions with progressive relationship are generated when contextual information is available. (3) More visual tokens retain more visual details. In sample (c), the corresponding AD becomes more detailed as the number of learnable vectors increases. Finally, we compare our generated ADs with other methods in sample (d). The description generated by \ours{} is more consistent with the video content than results of AutoAD-\uppercase\expandafter{\romannumeral2} and ClipCap. We find that result of MM-Narrator, which generates AD by prompting GPT-4, contains a lot of details that don’t actually exist. We speculate that this is because GPT-4 is trained on longer texts with a large number of detailed descriptions, which results in a serious hallucination problem.
\section{Conclusion}\label{sec:conclusion}

In this work, we present a simple and unified framework called \ours{} for Audio Description (AD) generation task by prompting pre-trained LLMs with interleaved multimodal sequence as input. Compared with previous work, our \ours{} is able to leverage more precise character information provided by the character-refinement module and fully utilize rich contextual information to generated ADs. \ours{} achieves the state-of-the-art performance on multiple AD generation benchmarks. We also conduct comprehensive ablation studies to validate the effectiveness of different components, which demonstrates that fine-grained feature alignment, precise character information, and contextual data can benefit AD generation. We hope our work could facilitate research in this community.

\noindent\textbf{Acknowledgements:} This work is supported by the National Key R$\&$D Program of China (No. 2022ZD0160900), ERC AdG Project (No. 101021347), Ant Group Research Intern Program, Jiangsu Frontier Technology Research and Development Program (No. BF2024076), and the Collaborative Innovation Center of Novel Software Technology and Industrialization.
\newpage
\appendix
\renewcommand\thesection{\Alph{section}}
\renewcommand\thefigure{S\arabic{figure}}
\renewcommand\thetable{S\arabic{table}}
\renewcommand\theequation{S\arabic{equation}}
\setcounter{figure}{0}
\setcounter{table}{0}
\setcounter{equation}{0}
\setcounter{page}{1}
\maketitlesupplementary

\section*{Appendix}

\section{Overview}\label{appendix:overview}

The supplementary material provides more implementation details, ablation analyses and qualitative results to show deep insights into our method. Specifically, ~\Cref{supp:imple} describes the architecture details of our model and experiment settings. ~\Cref{supp:ablation} provides more ablation studies on our \ours{}. To further show the effectiveness of our model design, we show more qualitative results of \ours{} in ~\Cref{supp:qualita}.

\section{Implementation Details}
\label{supp:imple}

\subsection{Architecture details}
We show the architecture details of our visual mapping network for MAD-eval-Named benchmark and character-refinement module in ~\cref{tab::arch}. Note that for CMDAD and TVAD datasets, we use the training set of CMDAD to train our character-refinement module and apply InternVideo2.5~\cite{internvideo2_5} to extract video frame features. The character-refinement performances are shown in ~\cref{tab::char}. Note that the character-refinement module for CMDAD and TVAD is trained on the training set of CMDAD and evaluated on both CMDAD and TVAD.

\begin{table}[h]
\caption{Architecture details of visual mapping network and character-refinement module in \ours{}.}\label{tab::arch}
\centering
\resizebox{\linewidth}{!}{
    \setlength{\tabcolsep}{6pt}
    \begin{tabular}{c|c|c}
\hline
\multirow{6}{*}{\shortstack{Visual mapping network \\ (GPT-2-B32)}}      & projection layer & 512$\rightarrow$768 \\
                                             & num latent       & 30 \\
                                             & num blocks       & 2 \\
                                             & channel          & 768 \\
                                             & num head         & 12 \\
                                             & ffn dimension    & 3072 \\ \hline
\multirow{6}{*}{\shortstack{Visual mapping network \\ (LLaMA-L14)}}      & projection layer & 768$\rightarrow$4096 \\
                                             & num latent       & 30 \\
                                             & num blocks       & 2 \\
                                             & channel          & 4096 \\
                                             & num head         & 32 \\
                                             & ffn dimension    & 16384 \\ \hline
\multirow{4}{*}{\shortstack{Character-refinement module \\ (for MAD-eval-Named)}} & 
                                               num blocks       & 3 \\
                                             & channel          & 768 \\
                                             & num head         & 12 \\
                                             & ffn dimension    & 3072 \\ \hline
\multirow{5}{*}{\shortstack{Character-refinement module \\ (for CMDAD\&TVAD)}} & 
projection layer & 4096$\rightarrow$4096 \\
                                             & num blocks       & 1 \\
                                             & channel          & 4096 \\
                                             & num head         & 32 \\
                                             & ffn dimension    & 16384 \\ \hline
\end{tabular}
}
\end{table}

\begin{table}[h]
\caption{Architecture details of visual mapping network and character-refinement module in \ours{}.}\label{tab::char}
\centering
\resizebox{0.65\linewidth}{!}{
    \setlength{\tabcolsep}{6pt}
\begin{tabular}{c|c|c}
\hline
Dataset        & Precision$\uparrow$ & Recall$\uparrow$ \\ \hline
MAD-eval-Named &     0.41      &    0.77    \\ \hline
CMDAD          &     0.26      &    0.94    \\ \hline
TVAD           &     0.27      &    0.94    \\ \hline
\end{tabular}
}
\vspace{-10pt}
\end{table}

\subsection{Datasets}

\noindent \textbf{MAD-Named}~\cite{MAD, autoad}: The MAD-Named benchmark consists of two parts: MAD-v2-Named for training and MAD-eval-Named for testing. Specifically, MAD-v2-Named contains 334,296 ADs and 628,613 subtitles collected from 488 movies, while MAD-eval-Named contains 6,520 ADs and 10,602 subtitles collected from 10 movies. Annotation includes the start and end time of each AD and the AD contents without any post-processing on character names. We notice there are many homophonic name mismatches in MAD-Named, such as an actor's name being 'Gray' in character information from IMDb but 'Grey' in the ad annotation. We thus processed these mismatched information to ensure that the same actor's name remains consistent.

\noindent \textbf{CMDAD}~\cite{Autoad3}: CMDAD is a movie AD dataset that contains 101k ADs for more than 1432 movies, with 100 movies split for evaluation.

\noindent \textbf{TVAD}~\cite{AutoAD0}: TVAD is a recently proposed TV-series AD dataset, which contains 31k ADs for training and 3k ADs for evaluation.

\noindent \textbf{AudioVault-AD}~\cite{autoad}: AudioVault-AD is a text-only dataset composed of 3.3 million AD utterances collected from 7,057 movies downloaded from the AudioVault website. Movies in AudioVault-AD are not included in the MAD dataset.

\subsection{Baselines}
In this paper, we compare our \ours{} with the following AD generation methods:

\noindent \textbf{ClipCap}~\cite{ClipCap}. The ClipCap model converts the CLIP~\cite{CLIP} feature of visual inputs into embeddings with a mapping network. Then the output embeddings will be used as prefix to prompt GPT-2~\cite{gpt2} to generate corresponding captions.

\noindent \textbf{AutoAD-\uppercase\expandafter{\romannumeral1}}~\cite{autoad}. AutoAD-\uppercase\expandafter{\romannumeral1} follows ClipCap and concatenate previous AD descriptions and subtitles in movie with visual embeddings to prompt the fronzen GPT-2~\cite{gpt2} for AD generation. This approach further apply partial-data pretrain to address the issue of insufficient AD data.

\noindent \textbf{AutoAD-\uppercase\expandafter{\romannumeral2}}~\cite{autoad2}. This method applies a Flamingo-style~\cite{Flamingo} architecture for AD generation and introduces an external Character Bank to enable their model to label characters appearing in the movie. AutoAD-\uppercase\expandafter{\romannumeral2} also presents an AD temporal proposal module to determine whether AD should be  inserted in the given pause in dialogue.

\noindent \textbf{AutoAD-\uppercase\expandafter{\romannumeral3}}~\cite{Autoad3}. AutoAD-\uppercase\expandafter{\romannumeral3} follows BLIP2~\cite{blip2} to use Q-former architecture to bridge the visual space with the language space. Then the model can generate textual outputs with a large language model. AutoAD-\uppercase\expandafter{\romannumeral3} also proposed a large-scale HowTo-AD dataset for pre-training.

\noindent \textbf{AutoAD-Zero}~\cite{AutoAD0}. AutoAD-Zero designs a pipeline for character recognition with face detection methods and prompts LLM by circling character faces. A two-stage training-free method is proposed for AD generation, which consists of (i) VLM-Based Video Description and (ii) LLM-Based AD Summary.

\noindent \textbf{MM-Narrater}~\cite{MM-Narrator}. MM-Narrater employs specialized vision and audio expert models to extract multimodal information from the input video clip. The outputs, along with movie subtitles and previous AD descriptions are used to build prompt to query GPT-4~\cite{GPT4} or GPT-4V~\cite{gpt4v} for AD generation. Besides, MM-Narrater utilizes retrieval enhancement and in context learning to improve the quality of generated AD. 

\subsection{Metircs}
In this paper, we use both classic captioning metrics and newly proposed metrics for evaluation. Classic captioning metrics include ROUGE-L~\cite{rouge} and
CIDEr~\cite{cider}. In this section, we mainly introduce newly proposed metrics: R@k/N, CRITIC and LLM-AD-eval.

\noindent \textbf{R@k/N}~\cite{autoad2}: R@k/N is a retrieval metric that distinguishes the predicted text among a set of neighbours. The parameters $k$ and $N$ mean within a temporal window of $N$ neighbouring reference ADs, whether the predicted AD can retrieve the corresponding reference AD at top-k position.

\noindent \textbf{CRITIC}~\cite{Autoad3}: CRITIC assesses the precision of character recognition in generated ADs. Specifically, a co-referencing model is utilized to substitute ambiguous pronouns in ADs with official names from the character banks. Subsequently, two sets of names from predicted and ground truth ADs are compared, and the IoU is computed to yield a CRITIC score.

\noindent \textbf{LLM-AD-eval}~\cite{Autoad3}: LLM-AD-eval utilises LLMs to judge the quality of generated ADs by scoring them between 1 (lowest) and 5 (highest). We use llama2-7b-chat~\cite{llama} for the evaluation in our experiments.

\section{More Ablation Studies}
\label{supp:ablation}

In this section, we explore more ablation studies on our \ours{}, which are not displayed in the main paper due to space limitation. All experiments are conducted with the character-refinement module and no pre-training is applied.

\subsection{Ablation on visual mapping network}
As stated in Sec. 3.2 of the main paper, there are multiple reasons why we choose a multi-layer transformer encoder with a fixed number of learnable vectors as our mapping network. Here we compare our visual mapping network with two different visual mapping designs: MLP and multi-layer transformer encoder without learnable vectors. Results in 
~\cref{tab::visual_map} show that no interaction between video frames (MLP as visual mapping network) gets the worst performance. Allowing interaction between video frames(transformer encoder as visual mapping network) brings better results, but the length of visual embeddings is limited to be consistent with the number of frames(8 in our experiments). 
Our visual mapping network with 30 learnable vectors performs the best.

\begin{table}[t]
\caption{\textbf{Ablation on the structure of visual mapping network.} \textit{latent} denotes the number of learnable vectors in our visual mapping network. Experiments are conducted with \ours{}(GPT-2-B32)}\label{tab::visual_map}
\centering
\resizebox{0.85\linewidth}{!}{
    \setlength{\tabcolsep}{6pt}
    \begin{tabular}{c|ccc}
\toprule
\textbf{Visual mapping network}     & \multicolumn{1}{c}{\textbf{RL$\uparrow$}}   & \multicolumn{1}{c}{\textbf{C$\uparrow$}}    & \multicolumn{1}{c}{\textbf{R@5/16$\uparrow$}} \\ \hline
MLP  & 14.0 &  20.2    &   45.5     \\
Transformer encoder  & 15.2 &  23.4    &  49.2     \\
Ours(latent=10)  & 15.4 &  22.4    & 49.0     \\
Ours(latent=30)  & \textbf{15.7} &  \textbf{23.7}    &   \textbf{49.4}     \\
\bottomrule
\end{tabular}
}

\end{table}

\begin{table}[t]
\caption{\textbf{Ablation on the impact of sharing visual mapping network}. \textit{Share?} shows whether we encode both video and image with one single visual mapping network.}\label{tab::share}
\centering
\resizebox{0.8\linewidth}{!}{
    \setlength{\tabcolsep}{6pt}
    \begin{tabular}{c|c|ccc}
\toprule
\textbf{Methods}   & \textbf{Share?} & \multicolumn{1}{c}{\textbf{RL$\uparrow$}}   & \multicolumn{1}{c}{\textbf{C$\uparrow$}}    & \multicolumn{1}{c}{\textbf{R@5/16$\uparrow$}} \\ \hline
\multirow{2}{*}{\shortstack{\ours{} \\ (GPT-2-B32)}}   &  \xmark      & \textbf{15.7}    & \textbf{23.7}  &    \textbf{49.4}    \\
&   \cmark       &   15.5  &  22.9 & 48.8  \\ \hline
\multirow{2}{*}{\shortstack{\ours{} \\ (LLaMA-L14)}} & \xmark  & \textbf{16.5}    & \textbf{25.9}  &    52.5    \\
&  \cmark       &  16.3   &  25.8 &  \textbf{53.6}  \\
 \bottomrule
\end{tabular}
}

\end{table}

\subsection{Ablation on sharing visual mapping network}
Since in \ours{}, the structure of video mapping network is the same as image mapping network, we in this section study the impact of using a single visual mapping network to encode both video and image. The results are shown in ~\cref{tab::share}. We can see that encoding video and image with two separate mapping networks is important to \ours{}(GPT-2-B32), while not necessary for \ours{}(LLaMA-L14). This reflects that when visual features and LLM are good enough, images and videos can be mixed together for training the mapping network.

\subsection{Ablation on image-video interaction}
In the main paper, we encode video and image into visual tokens and apply the frozen LLM for interaction between video and image. To study whether more interaction between image and video can benefit AD generation, we replace the input of visual mapping network as concatenation of image and video. Specifically, we replace the input with concatenation of current character's image and the video for image mapping network. For video mapping network, we replace the input with concatenation of all recognized characters and the video. In this way, we investigate whether the visual mapping network can extract better character and video representations by more interaction. The results are shown in ~\cref{tab::interaction}, which indicates that more interaction between image and video for visual mapping can not further benefit our \ours{}.

\begin{table}[t]
\caption{\textbf{Ablation on more interaction between image and video}. \textit{Inter-I.?} shows whether we concatenate character's image and the video clip as input of image mapping network. \textit{Inter-V.?} shows whether we concatenate all character images and the video clip as input of video mapping network.}\label{tab::interaction}
\centering
\resizebox{0.9\linewidth}{!}{
    \setlength{\tabcolsep}{6pt}
    \begin{tabular}{c|c|c|ccc}
\toprule
\textbf{Methods}   & \textbf{Inter-I.?} & \textbf{Inter-V.?} & \multicolumn{1}{c}{\textbf{RL$\uparrow$}}   & \multicolumn{1}{c}{\textbf{C$\uparrow$}}    & \multicolumn{1}{c}{\textbf{R@5/16$\uparrow$}} \\ \hline
\multirow{4}{*}{\shortstack{\ours{} \\ (GPT-2-B32)}}   &  \xmark  &  \xmark    & \textbf{15.7}    & 23.7  &    \textbf{49.4}    \\
  &  \xmark  &  \cmark    &  15.7   &  23.1   &  49.2  \\
&   \cmark   &  \xmark    &  15.7   & 23.6  & 49.1  \\
&   \cmark   &  \cmark    &  15.5   & \textbf{23.8}  & 49.1  \\\hline
\multirow{2}{*}{\shortstack{\ours{} \\ (LLaMA-L14)}} & \xmark & \xmark  & \textbf{16.5}  & \textbf{25.9}  &  52.5  \\
& \xmark  &  \cmark       &   16.5    & 25.4  &  \textbf{53.2}     \\
& \cmark  &  \xmark       &   16.5    & 25.7  &  52.2    \\
& \cmark  &  \cmark       &   16.3    & 25.0  &  52.5     \\
\bottomrule
\end{tabular}
}

\end{table}

\subsection{Ablation on the Threshold in Character-Refinement Module}
We conduct ablation study on the impact of threshold in Character-Refinement Module to our Uni-AD and the results in \cref{tab::threshold} show that the threshold has a considerable impact on AD generation. High threshold may lead to excessive loss of character information thus gets poor results.

\begin{table}[h]
\caption{\textbf{Ablation on the Threshold value in Character-Refinement Module}.}\label{tab::threshold}
\centering
\resizebox{\linewidth}{!}{
    \setlength{\tabcolsep}{6pt}
\begin{tabular}{cccc|cccc}
\hline
Threhold &  \textbf{RL$\uparrow$}   & \textbf{C$\uparrow$}    & \textbf{R@5/16$\uparrow$} & Threhold & \textbf{RL$\uparrow$}   & \textbf{C$\uparrow$}    & \textbf{R@5/16$\uparrow$} \\ \hline
0.3 & 16.5 & 25.7 & 52.4 & 0.5 & \textbf{16.8} & \textbf{27.3} & 53.3 \\
0.7 & 16.2 & 24.8 & \textbf{55.4} & 0.9 & 12.8 & 16.1 & 54.3 \\ \hline
\end{tabular}
}

\end{table}

\subsection{Limitation of contextual information and character-refinement module}
The core contribution of this paper is the proposal of generating ADs with interleaved multimodal sequence to achieve finer-grained modal alignment and better accomplish the AD narration task. The concurrent work AutoAD-III~\cite{Autoad3} also adopts this method to prompt large language models, while introducing the visual branch of VideoLLaMA~\cite{videollama} as a video encoder, achieving good results by leveraging pre-trained vision models. All these results demonstrate the feasibility of using interleaved multimodal sequence to prompt large language models for AD generation.

In addition to processing various inputs into a unified interleaved multimodal sequence, we also propose two sub-modules, contextual information modeling and character-refinement module. We have validated the effectiveness of these two modules on the main dataset MAD. For the recently proposed CMDAD and TVAD datasets, this section employs new experimental settings, i.e., using the visual branch of VideoLLaMA~\cite{videollama} as the video encoder and utilizing the character prediction results provided by AutoAD-Zero~\cite{AutoAD0}, conducting ablation experiments on the new datasets to verify the effectiveness of these two modules.

First, we perform an ablation study on the character-refinement module, and the results are shown in~\cref{tab::limit1}.  
It can be observed that after using the more accurate character prediction results provided by AutoAD-Zero~\cite{AutoAD0}, the character-refinement module leads to a decline in performance. The main reason is that the structure of the character-refinement module proposed in this paper is very simple, consisting of only a shallow multi-layer cross-attention model, and the amount of training data in the CMDAD dataset is relatively small, resulting in suboptimal training outcomes. Considering the current situation where the scarcity of training data is unlikely to change in the short term, future research could consider employing more powerful multimodal large models to achieve recognition of the main roles.

\begin{table}[t]
\caption{Abalation of contextual information and character-refinement module on CMDAD and TVAD. The results on TVAD here are obtained by directly evaluating the model after training on the CMDAD training set. The model is trained using the visual context information and contrastive loss function.}

\label{tab::limit1}
\centering
\begin{threeparttable}
  \resizebox{\linewidth}{!}{
  \begin{tabular}{c|c|ccc}
  \toprule
    \textbf{Char.?} & \textbf{Dataset} & \textbf{CIDEr$\uparrow$} &  \textbf{CRITIC$\uparrow$} & \textbf{LLM-AD-eval$\uparrow$} \\
    \hline
    \cmark & CMDAD & 21.8 &  41.9 & 2.92  \\
    \xmark & CMDAD & \textbf{22.4}  &  \textbf{42.2} & \textbf{2.93} \\ \hline
    \cmark & TVAD & 26.6  &  28.3 & 2.89  \\
    \xmark & TVAD & \textbf{26.9} &  \textbf{29.5} & \textbf{2.89} \\ \hline
  \end{tabular}
  }
  \end{threeparttable}
\vspace{-10pt}
\end{table}

We then conduct an ablation study on context information modeling using the character prediction results provided by AutoAD-Zero~\cite{AutoAD0}. The experiment here is performed only on CMDAD, as we only use the CMDAD training set for model training. The ablation results are shown in~\cref{tab:limit2}. It can be observed that with the new experimental setup, the benefits of introducing context information are not very significant: better result on the $Cider$ metric is achieved without using context information, while using context information leads to better performance on the $LLM\_AD\_eval$ metric, but the overall differences are not substantial. The main reason for this phenomenon may be that after introducing the powerful pre-trained visual encoder VideoLLaMA~\cite{videollama}, the model is already able to obtain better visual representations, thereby making the effect of context information modeling trained on the limited CMDAD data less noticeable.

\begin{table}[t]
\caption{Contextual information ablation on CMDAD. Results here are obtained using the character prediction results provided by AutoAD-Zero without the processing of the character-refinement module.}
\label{tab:limit2}
\centering
\begin{threeparttable}
\resizebox{\linewidth}{!}{
    \setlength{\tabcolsep}{6pt}
    \begin{tabular}{cc|ccc}
    \toprule
\textbf{Context-V}   & \textbf{C-Loss} & \textbf{CIDEr$\uparrow$} &  \textbf{CRITIC$\uparrow$} & \textbf{LLM-AD-eval$\uparrow$} \\ \hline
0  & \xmark      &  \textbf{22.6}   & 42.7  &  2.82  \\
1  &  \xmark     &  22.2   & \textbf{43.2}  &  2.92  \\
\hline
0 &   \cmark     &  22.3   & 42.4  &  2.85  \\
1 &   \cmark     &  22.4   & 42.2  &  \textbf{2.93}  \\
\hline
\end{tabular}
}
\end{threeparttable}
\vspace{-10pt}
\end{table}

\section{Additional Qualitative Analyses}
\label{supp:qualita}

\begin{figure}[b]
  \centering
  \includegraphics[width=0.45\textwidth]{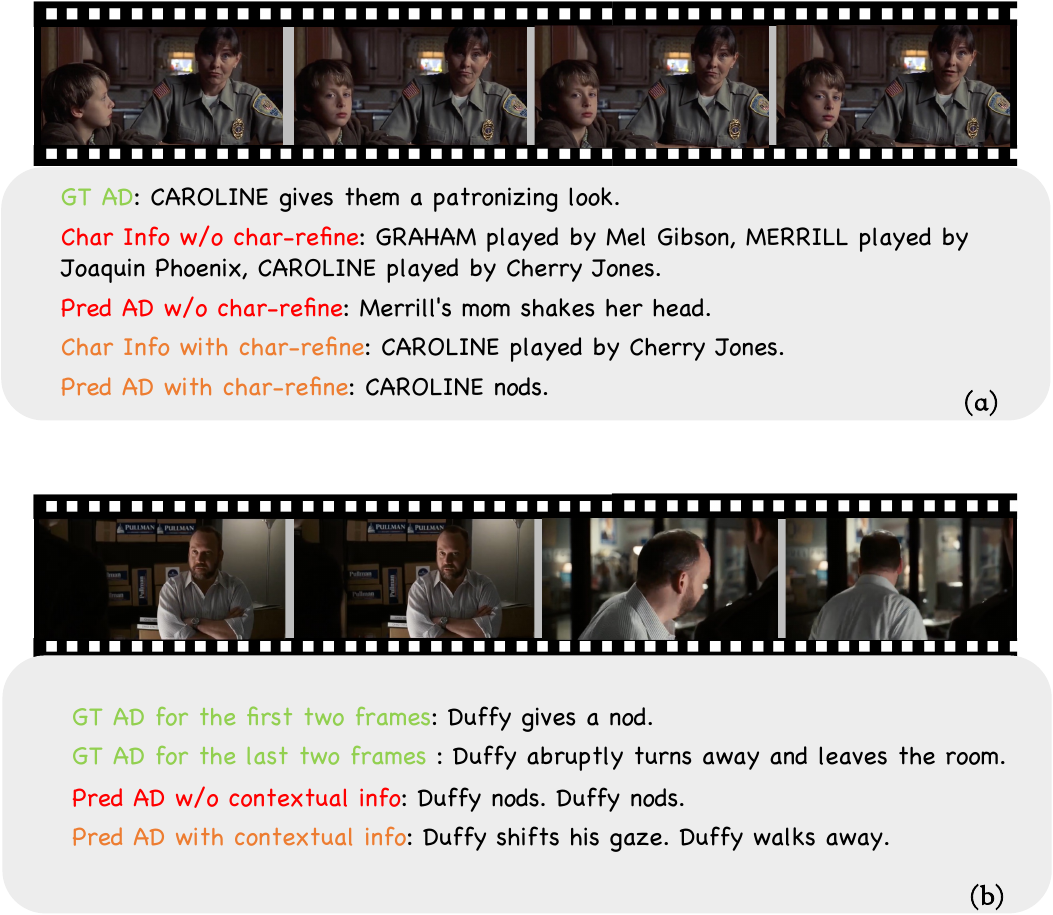}
  \caption{Qualitative analysis on character-refinement module and contextual information. Movies are selected from (a): Signs (2002), (b): The Ides of March (2011).}
  \label{fig:examples_supp1}
\end{figure}

\begin{figure}[t]
  \centering
  \includegraphics[width=0.45\textwidth]{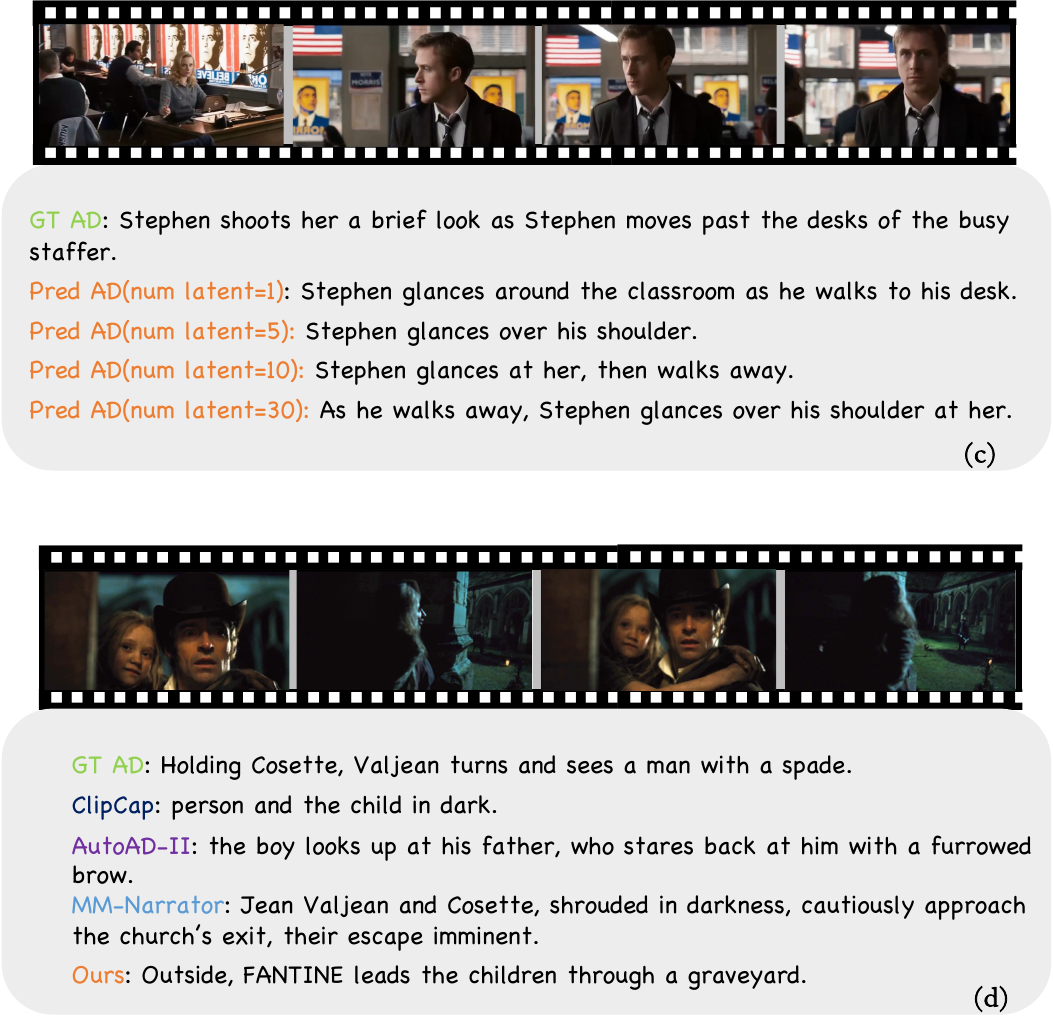}
  \caption{Qualitative analysis on number of learnable vectors and comparison with other approaches. Movies are selected from (c): The Ides of March (2011), (d): Les Mis$\acute{e}$rables (2012).}
  \label{fig:examples_supp2}
  \vspace{-10pt}
\end{figure}

~\Cref{fig:examples_supp1} and ~\Cref{fig:examples_supp2} shows more qualitative results of \ours{} on the MAD-eval dataset. Note that our character-refinement module can not only recognize AD-related characters, but also serve as a character information denoiser. For example, in sample (a) where characters \textit{Graham} and \textit{Merrill} do not appear in the video clip but are included in the character bank, our character-refinement module removes these noises and provides more precise character information. Though AD without character-refinement module also focuses on describing the female police officer, it can not figure out who she is since there are noises in the initial character bank, thus mistakes \textit{Caroline} as \textit{Merrill's} mom. In sample (c), we find that with more learnable vectors, our model can take the female character who appears at the beginning into AD generation. However, the female character is just ignored by our \ours{} with fewer learnable vectors.

\newpage
{
\small
\bibliographystyle{ieeenat_fullname}
\bibliography{main.bbl}
}

\end{document}